\documentclass[10pt,twocolumn,letterpaper]{article}

\usepackage{iccv}
\usepackage{times}
\usepackage{epsfig}
\usepackage{graphicx}
\usepackage{amsmath}
\usepackage{amssymb}

\usepackage[breaklinks=true,bookmarks=false]{hyperref}

\iccvfinalcopy 

\def\iccvPaperID{****} 

\ificcvfinal\fi

\begin{document}

\title{Mutual-GAN: Towards Unsupervised Cross-Weather Adaptation \\ with Mutual Information Constraint}

\author{Jiawei Chen,\,\,\,\,\,\, Yuexiang Li,\,\,\,\,\,\, Kai Ma,\,\,\,\,\,\, Yefeng Zheng\\
Tencent Jarvis Lab, Shenzhen, China\\
{\tt\small vicyxli@tencent.com}
}

\maketitle
\ificcvfinal\thispagestyle{empty}\fi

\begin{abstract}
   Convolutional neural network (CNN) have proven its success for semantic segmentation, which is a core task of emerging industrial applications such as autonomous driving. However, most progress in semantic segmentation of urban scenes is reported on standard scenarios, i.e., daytime scenes with favorable illumination conditions. In practical applications, the outdoor weather and illumination are changeable, e.g., cloudy and nighttime, which results in a significant drop of semantic segmentation accuracy of CNN only trained with daytime data. In this paper, we propose a novel generative adversarial network (namely Mutual-GAN) to alleviate the accuracy decline when daytime-trained neural network is applied to videos captured under adverse weather conditions. The proposed Mutual-GAN adopts mutual information constraint to preserve image-objects during cross-weather adaptation, which is an unsolved problem for most unsupervised image-to-image translation approaches (e.g., CycleGAN). The proposed Mutual-GAN is evaluated on two publicly available driving video datasets (i.e., CamVid and SYNTHIA). The experimental results demonstrate that our Mutual-GAN can yield visually plausible translated images and significantly improve the semantic segmentation accuracy of daytime-trained deep learning network while processing videos under challenging weathers.
\end{abstract}

\section{Introduction}

As one of the key techniques for autonomous driving, semantic segmentation of urban scenes gains increasing attentions from the community. Due to the development of convolutional neural networks (CNNs), semantic segmentation shows remarkable efficiency and reliability in the standard scenarios, \emph{e.g.,} daytime scenes with favorable illumination conditions. Yet, many existing systems have not been applied to practical applications, since they cannot deal with the adverse weathers such as nighttime and cloudy due to the lack of training data. Labeling more data under adverse weathers for network training may be one of the solutions to improve the semantic segmentation accuracy of CNNs. However, the pixel-wise annotation for semantic segmentation is laborious. Additionally, the illuminance of adverse conditions is often extremely weak that drastically deteriorates the structure, texture and color features of objects (\emph{e.g.,} road, car and pedestrian), which significantly increases the difficulty of accurate manual annotation.

\begin{figure}[!tb]
   \begin{center}
      \includegraphics[width=\linewidth]{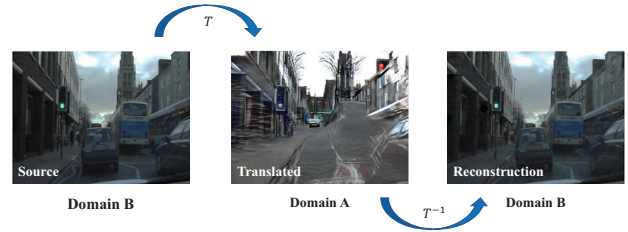}
   \end{center}
   \caption{Content distortion problem in the cloudy-to-sunny translation result produced by CycleGAN \cite{ZhuJ01}. The cycle-consistency is maintained with the bijective geometric transformation (\emph{e.g.,} translation, rotation, scaling, or even nonrigid transformation) and its inverse (\emph{i.e.,} $T$ and $T^{-1}$). The domain A and B contain images captured under sunny and cloudy weather, respectively.}
   \label{fig:distortion}
   \vspace{-3mm}
\end{figure}

One of the promising solutions is translating the images under adverse conditions to the standard scene (\emph{i.e.,} daytime) for accurate annotation/segmentation. Generative adversarial network (GAN) \cite{Goodf01}, which recently achieves a great success on generating high-quality synthetic images, becomes the mainstream for image translation. Many GAN-based image-to-image (I2I) translation approaches have been proposed, \emph{e.g.,} CycleGAN \cite{ZhuJ01}, UNIT \cite{LiuMY01}, and DRIT \cite{Lee01}. However, these GANs are frequently observed to yield the corruptions of image content in the translated images, which is unacceptable for the cross-weather translation task, due to the requirement of rigorous preservation of image-objects (\emph{e.g.,} car and pedestrian). To address the problem, Huang \emph{et al.} \cite{HuangS01} employed an additional segmentation to embed the semantic information to the generators, which enforced the CycleGAN to preserve the image-objects during daytime-nighttime translation. Nevertheless, the obvious drawback of this method is the demand of pixel-wise annotation of daytime and nighttime images.

In this paper, we propose a novel GAN, namely Mutual-GAN, for the cross-weather translation of urban scenes. A mutual information constraint, approximating the lower bound of the mutual information between source and translated images, is proposed to prevent image-objects from corruptions, which is not yet addressed by most GAN-based I2I approaches. Two publicly available driving video datasets (\emph{i.e.,} CamVid and SYNTHIA) are used to evaluate the effectiveness of our Mutual-GAN. Experimental results show that the proposed Mutual-GAN can not only produce plausible cross-weather images, while impeccably preserving the image-objects, but also significantly increase the semantic segmentation accuracy of daytime-trained networks (\emph{e.g.,} PSPNet \cite{Zhao_2017_CVPR}) under adverse conditions.

\section{Related Work}
The areas highly related to our study are unsupervised I2I translation and mutual information. In this section, we briefly review the previous works on these topics.

\subsection{Unsupervised Image-to-Image Translation}
Witnessing the success of cycle-consistency-based approaches (\emph{e.g.,} CycleGAN \cite{ZhuJ01}, DiscoGAN \cite{Kim01}, and DualGAN \cite{Yi01}), an increasing number of researchers \cite{shrivastava2017learning,Chen_2018_CVPR,Ma_2018_CVPR,fu2019geometry} made their efforts to the area of unpaired I2I translation. For example, UNIT \cite{LiuMY01}, a recently proposed model, assumes that there exists a shared-latent space in which a pair of corresponding images from different domains could be mapped to the same latent representation. Through such latent representation, the I2I translation can be achieved. To further increase the output diversity, Lee \emph{et al.} \cite{Lee01} proposed a disentangled representation framework, namely DRIT, with unpaired training data. DRIT embedded images into two spaces---a domain-invariant content space capturing shared information across domains, and a domain-specific attribute space to achieve diversity of the translated results. However, none of those approaches explicitly takes the image content preservation into account during translation, which may result in content distortion of the translated images.

\subsection{Domain Adaptation of Urban Scene} Generative adversarial networks have been widely used for the translation of urban scenes such as daytime-nighttime \cite{DaiD2018,HuangS01,SunL01}, synthetic-realistic \cite{Chen_2019_CVPR_2,Chen_2018_CVPR_2,Sankaranarayanan_2018_CVPR}, and cross-dataset \cite{Choi_2019_ICCV,Li_2019_CVPR} adaptation. In this paper, we focus on the cross-weather adaptation, \emph{i.e.,} translation between the standard (daytime) and adverse (nighttime and cloudy) scenes. A typical study in this area is AugGAN \cite{HuangS01}, which performed satisfactory daytime-nighttime translation to augment the amount of training data for neural networks. Nonetheless, AugGAN required pixel-wise annotations to overcome the aforementioned problem (\emph{i.e.,} image-object corruptions in the translated images), which were difficult to acquire and inapplicable for practical applications. 

To reduce the workload of manual annotation, we try to address the problem of content distortion in an unsupervised manner. Chen \emph{et al.} \cite{Jiaweichen2020} implemented a VideoGAN to maintain the style-consistency across a driving video during unsupervised video-to-video translation. Xie \emph{et al.} \cite{Xie_2020_ECCV} proposed an unsupervised cross-weather adaptation approach, namely OP-GAN, by integrating a self-supervised module into the CycleGAN. These two approaches are current state-of-the-art in the area, which are involved as the baseline in this study.

\subsection{Mutual Information}
Mutual information (MI) was first adopted to model the relationship of data from different domains by Shi and Sha \cite{ShiY2012}. They optimized the metric of MI to adapt classifiers trained on a labeled source domain to an unlabeled target domain. However, their computation of MI built upon the strong prior-knowledge, \emph{i.e.,} discriminative clustering, which is unable to be applied to neural networks. In more recent researches, the idea of MI has been introduced to improve the performance of GANs. InfoGAN \cite{NIPS2016_6399} was the first work in the area, which disentangled the representative features by maximizing the mutual information between a small subset of the latent variables and the observation. Na \emph{et al.} \cite{Na2019arxiv} proposed a mutual information loss (MILO), which optimized the same lower bound of MI, to improve the quality of synthetic images produced by the unsupervised I2I translation. Gholami \emph{et al.} \cite{GholamiTIP} proposed to optimize the Barber \& Agakov lower bound of MI for multi-target domain adaptation. In this paper, we apply MI to deal with the distortion problem occurred in most cycle-consistency-based GANs, which is the first work in the area of cross-weather translation, to our best knowledge.

\begin{figure*}[!htb]
   \begin{center}
      \includegraphics[width=\linewidth]{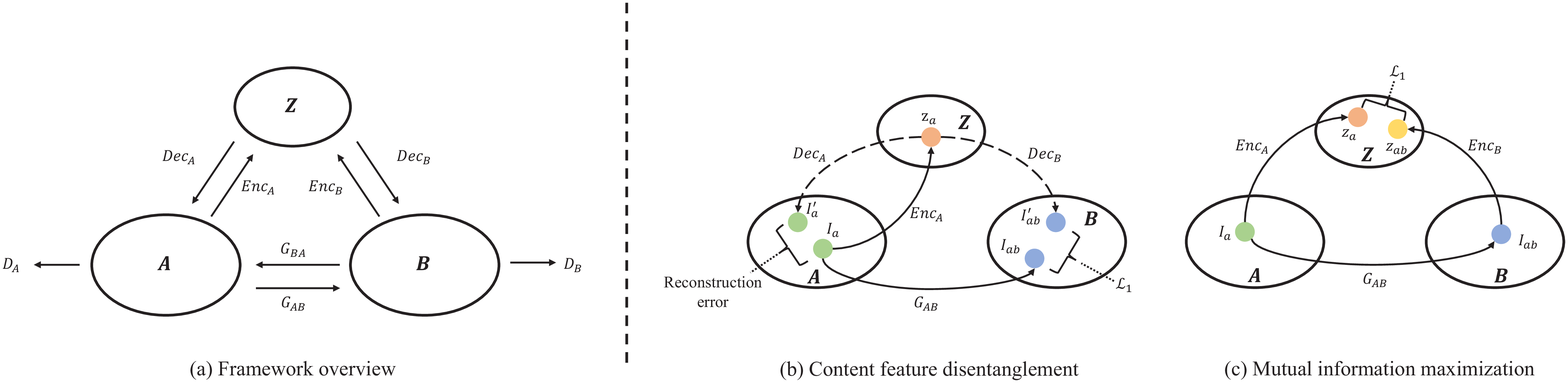}
      \caption{Network architecture (a) and two main stages---content feature disentanglement (b) and mutual information maximization (c)---of our Mutual-GAN.}
      \label{fig:pipeline}
   \end{center}
   \vspace{-3mm}
\end{figure*}

\section{Mutual-GAN}
Our Mutual-GAN aims to disentangle the content features from domain information for both the source and translated images, and then maximize the mutual information between the disentangled content features to preserve the image-objects. In this section, we first revisit the principle of CycleGAN and then present the modules for content feature disentanglement and mutual information maximization in details.

\subsection{Revisit Cycle-consistency-based GANs}
Cycle-consistency-based generative adversarial networks such as CycleGAN \cite{Zhang01} have two paired generator-discriminator modules, which are capable of learning two mappings, \emph{i.e.,} from domain A to domain B \{$G_{AB}$, $D_B$\} and vice versa \{$G_{BA}$, $D_A$\} . The generators ($G_{AB}$, $G_{BA}$) translate images between the source and target domains, while the discriminators ($D_A$, $D_B$) aim to distinguish the real and translated data. Thereby, the generators and discriminators are gradually updated during this adversarial competition.

The framework is usually supervised by two losses, \emph{i.e.,} adversarial loss ($\mathcal{L}_{adv}$) and cycle-consistency loss ($\mathcal{L}_{cyc}$). The adversarial loss encourages local realism of the translated data. Taking the translation from domain B to domain A as an example, the adversarial loss can be written as:
\begin{equation}
   \begin{aligned}
      \mathcal{L}_{adv}(G_{BA},D_{A})=\mathbb{E}_{x_{A}\sim p_{x_{A}}}\left [ (D_{A}(x_{A})-1)^{2} \right ] \\
      +\mathbb{E}_{x_{B}\sim p_{x_{B}}}\left [ (D_{A}(G_{BA}(x_{B})))^{2} \right ]
   \end{aligned}
\end{equation}
where $p_{x_{A}}$ and $p_{x_{B}}$ denote the sample distributions of domain A and B, respectively; $x_{A}$ and $x_{B}$ are samples from domain A and B, respectively.

The cycle-consistency loss ($\mathcal{L}_{cyc}$) tackles the requirement of paired training data. The idea behind the cycle-consistency loss is that the translated data from the target domain can be exactly converted back to the source domain, which can be expressed as:
\begin{equation}
   \begin{aligned}
      \mathcal{L}_{cyc}(G_{BA},G_{AB})=\mathbb{E}_{x_{A}\sim p_{x_{A}}}\left [ \left \| G_{BA}(G_{AB}(x_{A}))-x_{A} \right \|_{1} \right ] \\
      +\mathbb{E}_{x_{B}\sim p_{x_{B}}}\left [ \left \| G_{AB}(G_{BA}(x_{B}))-x_{B} \right \|_{1} \right ].
   \end{aligned}
\end{equation}

With these two losses, cycle-consistency-based GANs can perform image-to-image translation using unpaired training data. However, a recent study \cite{Zhang01} found that the cycle-consistency has an intrinsic ambiguity with respect to geometric transformations. Let $T$ be a bijective geometric transformation (\emph{e.g.,} translation, rotation, scaling, or even nonrigid transformation) with inverse transformation $T^{-1}$, the following generators $G^{'}_{AB}$ and $G^{'}_{BA}$ are also cycle consistent
\begin{equation}
   \begin{aligned}
      G^{'}_{AB} = G_{AB}T,\;G^{'}_{BA} = G_{BA}T^{-1}.
   \end{aligned}
\end{equation}

Consequently, due to the lack of penalty in content error between the source and translated images, an image translated from one domain to the other using cycle-consistent constraint may be geometrically distorted, as illustrated in Fig.~\ref{fig:distortion}. To address this problem, existing studies \cite{Zhang01,HuangS01} proposed to use segmentation sub-task with pixel-wise annotation, which is expensive and difficult to acquire for real applications, as an auxiliary regularization for the training of generators.

To loose the requirement of pixel-wise annotation, we propose to address the content distortion problem using a mutual information constraint. The pipeline of our approach (Mutual-GAN) is presented in Fig.~\ref{fig:pipeline} (a). Similar to current cycle-consistency-based GAN \cite{ZhuJ01}, our Mutual-GAN adopts paired generators ($G_{AB}$ and $G_{BA}$) and discriminators ($D_{B}$ and $D_{A}$) to achieve cross-domain image translation without paired training samples. Moreover, to preserve image contents during I2I domain adaptation without the pixel-wise annotation, the proposed Mutual-GAN disentangles the content features from domain information in an unsupervised manner using pairs of content feature encoders (\emph{i.e.,} $Enc_A$ and  $Enc_B$) and domain feature decoders (\emph{i.e.,}  $Dec_A$ and  $Dec_B$).

Specifically, The encoders (\emph{i.e.,} $Enc_A$ and  $Enc_B$) are responsible to embed the content information of source and translated images into the same latent space $Z$, while the decoders (\emph{i.e.,}  $Dec_A$ and  $Dec_B$) aim to transform the embedded content features to their own domains using domain-related information. Therefore, to alleviate the content distortion problem during image translation, we only need to maximize the mutual information between the content feature of source image and the translated image, which is achieved by our mutual information constraint.

\subsection{Content Feature Disentanglement}
The proposed Mutual-GAN utilizes the content feature encoders ($Enc_A$, $Enc_B$) and domain feature decoders ($Dec_A$, $Dec_B$) to disentangle the content features (\emph{i.e.,} image-objects) from domain information (\emph{i.e.,} weather and illumination). Since the mappings between domains A and B are symmetrical, we take the content feature distillation of images from domain A as an example, as shown in Fig.~\ref{fig:pipeline} (b). Given an input image ($I_a$), the encoder ($Enc_A$) embeds it into a latent space, which can be formulated as:
\begin{equation}
   \begin{aligned}
      z_a = Enc_A(I_a).
   \end{aligned}
\end{equation}

The embedded feature $z_a$ contains the information of content and domain A. To disentangle them, $z_a$ is mapped to domain A and B via $Dec_A$ and $Dec_B$, respectively
\begin{equation}
   \begin{aligned}
      I_a' = Dec_A(z_a), \;\;\; I_{ab}' = Dec_B(z_a)
   \end{aligned}
\end{equation}
where $I_{a}'$ and $I_{ab}'$ are the mapping results of $z_a$ to domain A and B, respectively.

Similar to Auto-Encoder, $I_{a}'$ can be seen as the reconstruction of $I_{A}$. As shown in Fig.~\ref{fig:pipeline}, apart from the X-shape dual Auto-Encoders, there is another translation path between domain A and B:
\begin{equation}
   \begin{aligned}
      I_{ab} = G_{AB}(I_a)
   \end{aligned}
\end{equation}
where $I_{ab}$ is the translated image yielded by $G_{AB}$.

Through simultaneously minimizing the pixel-wise $\mathcal{L}_1$ norm between $I_{ab}$ and $I_{ab}'$, and reconstruction error between $I_a$ and $I_{a'}$, $Dec_A$ and $Dec_B$ are encouraged to recover domain-related information from the latent space (in short, the encoders remove domain information and the decoders recover it), which enables them to map the $z_a$ to two different domains. Therefore, the information contained in $z_a$ is highly related to the image-objects without domain bias. The content feature distillation loss ($\mathcal{L}_{dis}$), combining aforementioned two terms, can be formulated as:
\begin{equation}
   \begin{aligned}
      \mathcal{L}_{dis} = ||I_{ab}-I_{ab}'||_1 + ||I_a - I_{a'}||_1.
   \end{aligned}
\end{equation}

\subsection{Mutual Information Maximization}
Using the content feature encoder and domain feature decoders, the content feature of source image $I_a$ can be embedded to $z_a$. To preserve image-objects during I2I translation, the translated image ($I_{ab}$) is enforced to contain similar information to the embedded content feature ($z_a$), which can be achieved by maximizing the mutual information ($\mathcal{I}$) between $I_{ab}$ and $z_a$. Therefore, we can formulate the mutual information loss ($\mathcal{L}_{MI}$) as: 
\begin{equation}
   \begin{aligned}
      \mathcal{L}_{MI} & = \mathcal{I}(z_a; I_{ab}) = H(z_a) - H(z_a|I_{ab})                                                                                                                     \\
                       & =\mathbb{E}_{x \sim I_{ab}}\left[\mathbb{E}_{z_{a}^{\prime} \sim P\left(z_{a} | x\right)}\left[\log P\left(z_{a}^{\prime} | x\right)\right]\right]+H\left(z_{a}\right).
   \end{aligned}
\end{equation}

However, according to \cite{NIPS2016_6399}, the posterior probability $P(z_a|x)$ is difficult to be directly estimated. To this end, we utilize an auxiliary distribution $Q(.)$, which is easily accessed, to approximate the posterior probability:
\begin{equation}
   \begin{aligned}
       & \mathcal{I}\left(z_{a}; I_{ab}\right)=\mathbb{E}_{x \sim I_{ab}}\left[\mathbb{E}_{z_{a}^{\prime} \sim P\left(z_{a} | x\right)}\left[\log P\left(z_{a}^{\prime} | x\right)\right]\right]+H\left(z_{a}\right) \\
       & =\mathbb{E}_{x \sim I_{ab}}\left[KL(P(\cdot | x) \| Q(\cdot | x))+\mathbb{E}_{z_{a}^{\prime} \sim P\left(z_{a} | x\right)}\left[\log Q\left(z_{a}^{\prime} | x\right)\right]\right]                         \\
       & \,\,\,\,\,\,+H\left(z_{a}\right)
   \end{aligned}
\end{equation}
where $KL$ is the KL-divergence.

Since $KL(P(\cdot | x) \| Q(\cdot | x)) \geq 0$, we can obtain the lower bounded mutual information based on variational information maximization \cite{NIPS2003}:
\begin{equation}
   \begin{aligned}
      \mathcal{I}\left(z_{a}; I_{ab}\right) \geq \mathbb{E}_{x \sim I_{ab}}\left[\mathbb{E}_{z_{a}^{\prime} \sim P\left(z_{a} | x\right)}\left[\log Q\left(z_{a}^{\prime} | x\right)\right]\right]+H\left(z_{a}\right)
   \end{aligned}.
\end{equation}

Hence, the proposed Mutual-GAN can maximize the $\mathcal{I}(z_a;I_{ab})$ via optimizing the lower bound---the lower bound becomes tight as the auxiliary distribution $Q$ approaches the true posterior distribution: $\mathbb{E}_{x}\left[KL(P(\cdot | x) \| Q(\cdot | x))\right] \rightarrow 0$.

In practice, $\mathbb{E}_{x \sim I_{ab}}[Q\left(z_{a}^{\prime} | x\right)]$ can be parametrized using $Enc_B$: $z_{a}^{\prime} \rightarrow z_{ab}$, as illustrated in Fig.~\ref{fig:pipeline} (c). Therefore, the maximization of $\mathcal{I}(z_a;I_{ab})$ can be achieved by aligning the distributions of $P(z_a|x)$ and $Q(z_{ab}|x)$ with $\mathbb{E}_{x\sim I_{ab}}$. In our experiments, the $\mathcal{L}_1$ norm is adopted for the distribution alignment:
\begin{equation}
   \begin{aligned}
      \mathcal{L}_{MI} = ||z_a - z_{ab}||_1.
   \end{aligned}
\end{equation}

\subsection{Objective}
With the previously defined feature distillation loss ($\mathcal{L}_{dis}$) and mutual information constraint ($\mathcal{L}_{MI}$), the full objective $\mathcal{L}$ for the proposed Mutual-GAN is summarized as:
\begin{equation}
   \begin{aligned}
      \mathcal{L} & = \mathcal{L}_{adv}\left(G_{BA},\ D_A\right)+\mathcal{L}_{adv}\left(G_{AB},\ D_B\right)   \\
                  & +\alpha \mathcal{L}_{cyc}\left(G_{AB}, \ G_{BA}\right)                                    \\
                  & + \mathcal{L}_{dis}(G_{AB}, Enc_{A}, Dec_{A}, Dec_{B})                                    \\
                  & + \mathcal{L}_{dis}(G_{BA}, Enc_{B}, Dec_{B}, Dec_{A})                                    \\
                  & + \mathcal{L}_{MI}(G_{AB}, Enc_{A}, Enc_{B}) + \mathcal{L}_{MI}(G_{BA}, Enc_{A}, Enc_{B})
   \end{aligned}\label{eq:objective}
\end{equation}
where $\mathcal{L}_{adv}$ and $\mathcal{L}_{cyc}$ are adversarial and cycle-consistency losses respectively, the same as that proposed in \cite{ZhuJ01}.

The optimization of $\mathcal{L}_{dis}$ and $\mathcal{L}_{MI}$ is performed in the same manner of $\mathcal{L}_{adv}$---fixing $Enc_A$/$Enc_B$, $Dec_A$/$Dec_B$, and $D_{A}$/$D_{B}$ to optimize $G_{BA}$/$G_{AB}$ first, and then optimize $Enc_A$/$Enc_B$, $Dec_A$/$Dec_B$, and $D_{A}$/$D_{B}$ respectively, with $G_{BA}$/$G_{AB}$ fixed. Therefore, similar to discriminators, our content encoders and domain decoders can directly pass the knowledge of image-objects to the generators, which helps them to improve the quality of translated results in terms of object preservation.

\subsection{Comparison with CycleGAN and InfoGAN}
We notice that our Mutual-GAN is highly related to CycleGAN \cite{ZhuJ01} and InfoGAN \cite{NIPS2016_6399}; therefore, it is worthwhile to emphasize the differences between the proposed model and the existing frameworks.

Compared to CycleGAN, our Mutual-GAN consists of two novel components: content feature disentanglement and mutual information maximization modules, which alleviate the problem of image-object distortion via maximizing the mutual information between content features ($z_a$ and $z_b$) distilled by the content feature disentanglement module. In comparison with InfoGAN, we propose a new paradigm to measure the mutual information (MI) between a feature ($z_a$) and an image ($I_{ab}$), since the proposed Mutual-GAN involves more network components for the MI calculation ($Enc_A$, $G_{AB}$, $Enc_B$ \emph{vs.} generator-only in InfoGAN).

\section{Experiments}
As aforementioned, semantic segmentation networks trained with standard scenario images (domain {\bf A}), \emph{i.e.,} daytime scenes with favorable illuminations, suffer from performance degradation while testing on real images under changeable weathers (domain {\bf B}), \emph{e.g.,} cloudy and nighttime. Hence, our goal is to narrow down the gap between images under different weathers not only in terms of visual perception \emph{i.e.,} plausible translated results, but also the representation in feature space, \emph{i.e.,} improvement of the robustness of models. We present the image-to-image translation results to evaluate the former factor. For the latter one, we evaluate the Mutual-GAN in the similar transfer learning scenario\footnote{The transfer learning scenario is more practical for realistic applications, as the well-trained deep learning networks can be directly applied to unlabeled data from another domain. In contrast, the data augmentation scenario still needs the pixel-wise annotation to re-train the networks.} to \cite{SunL01}.


\subsection{Datasets}
Experiments are conducted on two publicly available datasets to demonstrate the effectiveness of our Mutual-GAN. The datasets are:

\paragraph{\bf CamVid.} The dataset\footnote{http://mi.eng.cam.ac.uk/research/projects/VideoRec/CamVid/} \cite{BrostowSFC:ECCV08} contains driving videos under different weathers, \emph{e.g.,} cloudy and sunny. The task translating cloudy videos to the sunny domain is very challenging, as the cloudy videos are often very dark, which lose much detailed information. We conduct experiments on cloudy-to-sunny translation to evaluate our Mutual-GAN.

\paragraph{\bf SYNTHIA.} The dataset\footnote{http://synthia-dataset.net/} \cite{bengar2019temporal} consists of photo-realistic frames rendered from a virtual city. The night-to-day translation is a more difficult task than the cloudy-to-sunny, since the night domain suffers from severe loss of context information. We examine how the proposed Mutual-GAN performs on the night-to-day translation task using two sub-sequences (\emph{i.e.,} summer (daytime) and night (nighttime)) from the Highway, which is a subset of SYNTHIA.

\begin{figure*}[!htb]
   \begin{center}
      \includegraphics[width=\linewidth]{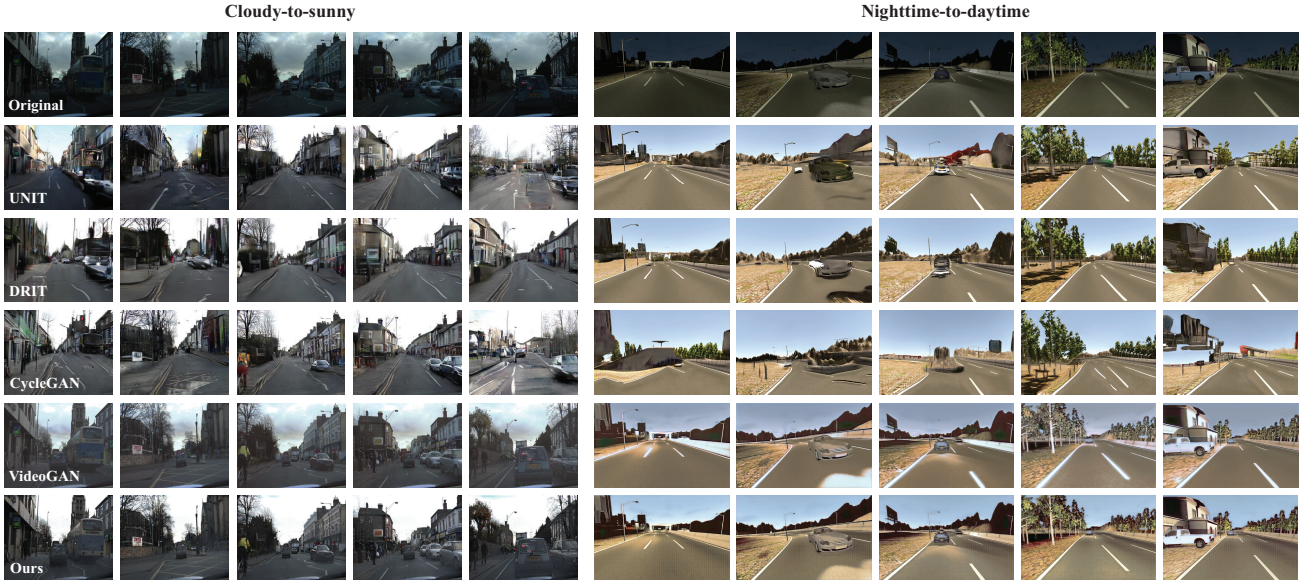}
      \caption{Comparison of cross-weather adaptation results yielded by different approaches. The image adaptation performance is evaluated on two tasks (cloudy-to-sunny and nighttime-to-daytime). The original images, translated results produced by UNIT \cite{LiuMY01}, DRIT \cite{Lee01}, CycleGAN \cite{ZhuJ01}, VideoGAN \cite{Jiaweichen2020} and our Mutual-GAN are presented. Since we do not reproduce the OP-GAN \cite{Xie_2020_ECCV}, its translation results are excluded from the figure.}
      \label{fig:visual_comparison}
   \end{center}
   \vspace{-3mm}
\end{figure*}

\begin{table*}[!tb]
   \centering
   \caption{Semantic segmentation IoU (\%) of the cloudy images from CamVid with different image-to-image translation frameworks.}\label{table_camvid_quantity}
   \footnotesize
   \begin{tabular}{l|ccccccccccccc}
      \hline
      {}                               & Bicyclist   & Building    & Car         & Pole        & Fence       & Pedestrian & Road      & Sidewalk  & Sign      & Sky         & Tree      & \bf mIoU    \\\hline\hline
      {\bf Sunny (validation)}         & 85.30       & 85.64       & 90.40       & 20.60       & 76.00       & 61.82      & 94.46     & 89.48     & 12.12     & 94.20       & 91.83     & 71.19       \\\hline\hline
      \multicolumn{12}{l}{\bf Cloudy (test)}                                                                                                                                                          \\\hline
      {Direct transfer}                & 31.76       & 51.35       & 38.97       & 15.09       & 10.55       & 32.58      & 61.82     & {51.68}   & 15.36     & 68.60       & 69.68     & 39.48       \\\hline
      {UNIT \cite{LiuMY01}}            & 0.67        & 49.28       & 6.99        & 6.10        & 1.54        & 0.79       & 45.05     & 15.16     & 0.00      & 62.61       & 39.34     & 19.94       \\\hline
      {DRIT \cite{Lee01}}              & 0.34        & 40.00       & 0.31        & 0.33        & 0.83        & 0.23       & 48.27     & 26.81     & 0.00      & 64.85       & 23.73     & 18.20       \\\hline
      {CycleGAN \cite{ZhuJ01}}         & 6.16        & 56.64       & 10.76       & 8.61        & 0.01        & 4.06       & 50.49     & 30.21     & 8.67      & 75.97       & 45.62     & 26.26       \\\hline
      {VideoGAN \cite{Jiaweichen2020}} & 33.57       & \bf 73.24       & 43.08       & 22.79       & 12.39       & 34.70      & 68.83     & 48.65     & \bf 22.09 & 76.38       & \bf 73.10 & 46.37       \\\hline
      {OP-GAN \cite{Xie_2020_ECCV}}    & {\bf 51.28} & {73.10} & {\bf 74.19} & 25.84       & 12.42       & \bf 42.75  & 70.48     & 51.74     & 14.71     & 81.09       & 72.40     & 51.40       \\\hline
      {Mutual-GAN (Ours)}              & 48.15       & 69.51       & 60.72       & {\bf 25.95} & {\bf 15.04} & \bf 42.75  & \bf 81.44 & \bf 69.49 & 10.84     & {\bf 91.78} & {70.63}   & {\bf 51.93} \\\hline\hline
      {Ours w/o $\mathcal{L}_{MI}$}    & {31.51}     & {67.12}     & {14.17}     & {18.85}     & {6.73}      & 23.54      & 73.49     & 59.33     & 19.05     & {87.05}     & {73.33}   & {41.82}     \\\hline
   \end{tabular}
   \vspace{-3mm}
\end{table*}

\subsection{Experiment Settings}

\paragraph{\bf Evaluation criterion.} The mean of class-wise intersection over union (mIoU) \cite{Zhao_2017_CVPR} is used to evaluate the improvement achieved by our Mutual-GAN on the semantic segmentation tasks for CamVid and SYNTHIA datasets.

\paragraph{\bf Baselines overview.} Several unpaired image-to-image domain adaptation frameworks, including CycleGAN \cite{ZhuJ01}, UNIT \cite{LiuMY01} and DRIT \cite{Lee01} are taken as baselines for the performance evaluation. The state-of-the-art frameworks (\emph{i.e.,} VideoGAN \cite{Jiaweichen2020} and OP-GAN \cite{Xie_2020_ECCV}) and the direct transfer approach, which directly takes the source domain data for testing without any adaptation, are also involved for comparison.\footnote{Since VideoGAN \cite{Jiaweichen2020} is only evaluated on CamVid, we reimplement the framework and report the performance of our reproduction. Instead, OP-GAN has been evaluated on both datasets adopted in this study; hence, we involve the accuracy reported in \cite{Xie_2020_ECCV} for comparison.} Note that the recently proposed GAN for image-based domain adaptation, \emph{i.e.,} AugGAN \cite{HuangS01}, is not involved for comparison, due to the strong prior-knowledge (pixel-wise annotation) used in the approach.

The benchmarking algorithms adopted in this study is the same to the previous studies \cite{HuangS01,Jiaweichen2020,Xie_2020_ECCV}. Some of the existing unsupervised domain adaptation (UDA) methods \cite{Li_2019_CVPR,Chang_2019_CVPR,NIPS2019_8335,Vu_2019_CVPR,Zou_2019_ICCV} are not involved for comparison, since they are feature-alignment (FA) based, while our Mutual-GAN is an image-to-image-based domain adaptation method. The difference between these two types of UDA methods is I2I-based methods are task-agnostic, compared to the FA-based domain adaptation. In other words, the FA-based methods require specific-annotations (\emph{e.g.,} pixel-wise annotation) of source domain to perform adaptive domain adaptation, which is not a necessary requirement for I2I-based approaches. To this end, it is difficult to fairly compare the performances between these two types of UDA approaches.

\paragraph{\bf Training details.} The proposed Mutual-GAN is implemented using PyTorch. The generator, discriminator, and Siamese network are iteratively trained for 200 epochs with the Adam solver \cite{kingma2014adam}. The split of training, validation and test sets for semantic segmentation on both CamVid and SYNTHIA datasets is consistent to \cite{Xie_2020_ECCV}. The baselines involved in this study adopt the same training protocol.

\subsection{Visualization of Translation Results}
The translation results for the two tasks generated by different image-to-image translation frameworks are presented in Fig.~\ref{fig:visual_comparison}, which illustrate the main problem of existing approaches (UNIT \cite{LiuMY01}, DRIT \cite{Lee01} and CycleGAN \cite{ZhuJ01})---image content corruptions. Due to the lack of penalty in content error between the source and translated images, the existing translation frameworks intend to overly edit the image content such as changing the shape and colors of image-objects, referring to the road and building in the CamVid and SYNTHIA translated images. This drawback restricts applications of the existing frameworks to the domain adaptation of urban scenes, which requires rigorous preservation of image-objects. The VideoGAN \cite{Jiaweichen2020} can maintain the image-objects, but fail to translate the image-weather. The clouds are clearly observed in its cloudy-to-sunny translated results. Oppositely, the proposed Mutual-GAN can excellently perform cross-weather translation, while preserving the image-objects.

\begin{table*}[!tb]
   \centering
   \caption{Semantic segmentation IoU (\%) of the night images from SYNTHIA with different image-to-image translation frameworks. (Veg. -- Vegetation, l.-m. -- lane-marking)}\label{table_synthia_quantity}
   \begin{tabular}{l|ccccccccccc}
      \hline
      {}                               & Sky       & Building  & Road        & Fence       & Veg.      & Pole        & Car       & Sign      & l.-m.       & \bf mIoU    \\\hline\hline
      {\bf Day (validation)}           & 94.20     & 80.83     & 97.76       & 89.34       & 67.61     & 40.42       & 66.98     & 2.41      & 80.14       & 70.06       \\\hline\hline
      \multicolumn{11}{l}{\bf Night (test)}                                                                                                                              \\\hline
      {Direct transfer}                & 25.67     & 58.49     & 89.16       & 64.44       & \bf 57.53 & 39.32       & 10.20     & 23.13     & 40.95       & 47.36       \\\hline
      {UNIT \cite{LiuMY01}}            & 91.93     & \bf 77.02 & 90.63       & 29.12       & 55.84     & 39.84       & \bf 74.34 & 28.16     & 51.19       & 61.21       \\\hline
      {DRIT \cite{Lee01}}              & 77.25     & 39.12     & 72.28       & 4.33        & 43.47     & 25.80       & 36.57     & 6.22      & 0.08        & 35.01       \\\hline
      {CycleGAN \cite{ZhuJ01}}         & 63.72     & 22.21     & 59.05       & 0.89        & 24.63     & 15.75       & 2.36      & 0.00      & 11.55       & 22.51       \\\hline
      {VideoGAN \cite{Jiaweichen2020}} & 91.59     & 43.29     & 88.78       & 49.56       & 39.59     & 33.84       & 45.24     & 15.11     & 67.47       & 54.67       \\\hline
      {OP-GAN \cite{Xie_2020_ECCV}}    & 21.90     & 66.22     & 86.78       & 7.7         & 54.86     & 39.11       & 85.09     & \bf 31.40 & 47.61       & 48.97       \\\hline
      {Mutual-GAN (Ours)}              & \bf 94.67 & 70.54     & {\bf 92.23} & {\bf 69.73} & 54.64     & {\bf 41.22} & {69.05}   & {30.29}   & {\bf 71.45} & {\bf 67.27} \\\hline\hline
      {Ours w/o $\mathcal{L}_{MI}$}    & 89.53     & 48.91     & 91.11       & 41.28       & 56.29     & 34.32       & 52.50     & 10.88     & 68.51       & 56.57       \\\hline
   \end{tabular}
   \vspace{-3mm}
\end{table*}

\subsection{Comparison with State-of-the-art}
In this section, we evaluate the performance of our Mutual-GAN on the two I2I domain adaptation tasks and compare with the state-of-the-art approaches, \emph{e.g.,} VideoGAN \cite{Jiaweichen2020} and OP-GAN \cite{Xie_2020_ECCV}.

\paragraph{\bf Cloudy-to-sunny adaptation on CamVid.}
The CamVid dataset contains four sunny videos (577 frames in total) and one cloudy video (124 frames). Each frame of the videos is manually annotated, which associates each pixel with one of the 32 semantic classes. Based on the widely-accepted protocol \cite{Zhu_2019_CVPR}, we focus on 11 classes including bicyclist, building, car, pole, fence, pedestrian, road, sidewalk, sign, sky, and tree. To evaluate the translation performance yielded by our Mutual-GAN, a semantic segmentation network (PSPNet\footnote{The top-1 solution (without extra training data) on the leaderboard of semantic segmentation on CamVid: https://paperswithcode.com/sota/semantic-segmentation-on-camvid.} \cite{Zhao_2017_CVPR}) is trained with the sunny frames and tested on the original cloudy frames and the translated ones. In the experiment, the sunny frames are separated to training (three videos) and validation (one video) sets. The evaluation results are shown in Table~\ref{table_camvid_quantity}.

Due to the loss of detailed information, it can be observed from Table~\ref{table_camvid_quantity} that the performance of PSPNet trained with sunny images dramatically drops to $39.48\%$ while tested on the original cloudy images. As the existing I2I domain adaptation approaches may encounter the content distortion problem and corrupt the image-objects, the semantic segmentation mIoU of PSPNet further degrades to $26.26\%$, $19.94\%$ and $18.20\%$ using the CycleGAN, UNIT and DRIT, respectively. The video-to-video domain adaptation approach (VideoGAN) marginally improves the mIoU to $46.37\%$, compared to the direct transfer. In contrast, the proposed Mutual-GAN significantly boosts the accuracy of most classes, especially for the sky (\emph{i.e.,} $+23.18\%$). The mIoU of our Mutual-GAN (\emph{i.e.,} $51.93\%$) is slightly higher than the state-of-the-art OP-GAN ($51.40\%$), which demonstrates its effectiveness on image-object preservation.



\paragraph{\bf Nighttime-to-daytime adaptation on SYNTHIA.}
We adopt two sub-sequences (summer (daytime) and night (nighttime)) from SYNTHIA to perform nighttime-to-daytime adaptation. The daytime and nighttime sequences contain 603 and 461 frames, respectively. SYNTHIA dataset provides pixel-wise semantic annotations for each frame, which can be categorized to nine semantic classes, as listed in Table~\ref{table_synthia_quantity}. The partition of training, validation and test sets complies with the same protocol to that of the CamVid dataset---the daytime images are separated to training and validation sets according to the ratio of 70:30, while all the nighttime images are used for test. The fully convolutional network (PSPNet \cite{Zhao_2017_CVPR}) is also adopted in this experiment to perform semantic segmentation.

The semantic segmentation mIoUs yielded by different testing strategies are shown in Table~\ref{table_synthia_quantity}. Similar to the cloudy images, the PSPNet trained with daytime images fails to properly process the nighttime images---an mIoU of $47.36\%$, due to the loss of information. The nighttime-to-daytime translation on SYNTHIA is a more challenging task compared to the cloudy-to-sunny translation on CamVid, since large areas of night images are dark, where the image-objects (\emph{e.g.,} cars) are difficult to recognize. Refer to Fig.~\ref{fig:visual_comparison}, most of existing image-to-image translation frameworks often create contents to fill the extremely dark areas, which consequently corrupts the original image-objects, \emph{e.g.,} cars and buildings. Due to these distortions, the images translated by CycleGAN and DRIT further decrease the mIoU of PSPNet to $22.51\%$ and $35.01\%$, respectively. The OP-GAN, VideoGAN and UNIT yield improvements to mIoU, \emph{i.e.,} $+1.61\%$, $+7.31\%$ and $+13.85\%$, respectively. Since our Mutual-GAN excellently prevents image-object corruptions during nighttime-to-daytime translation (refer to Fig.~\ref{fig:visual_comparison}), it achieves the best mIoU ($67.27\%$) for nighttime images, which is comparable to the validation result (70.06\%) tested on daytime images.

\begin{figure*}[!tb]
   \begin{center}
      \includegraphics[width=\linewidth]{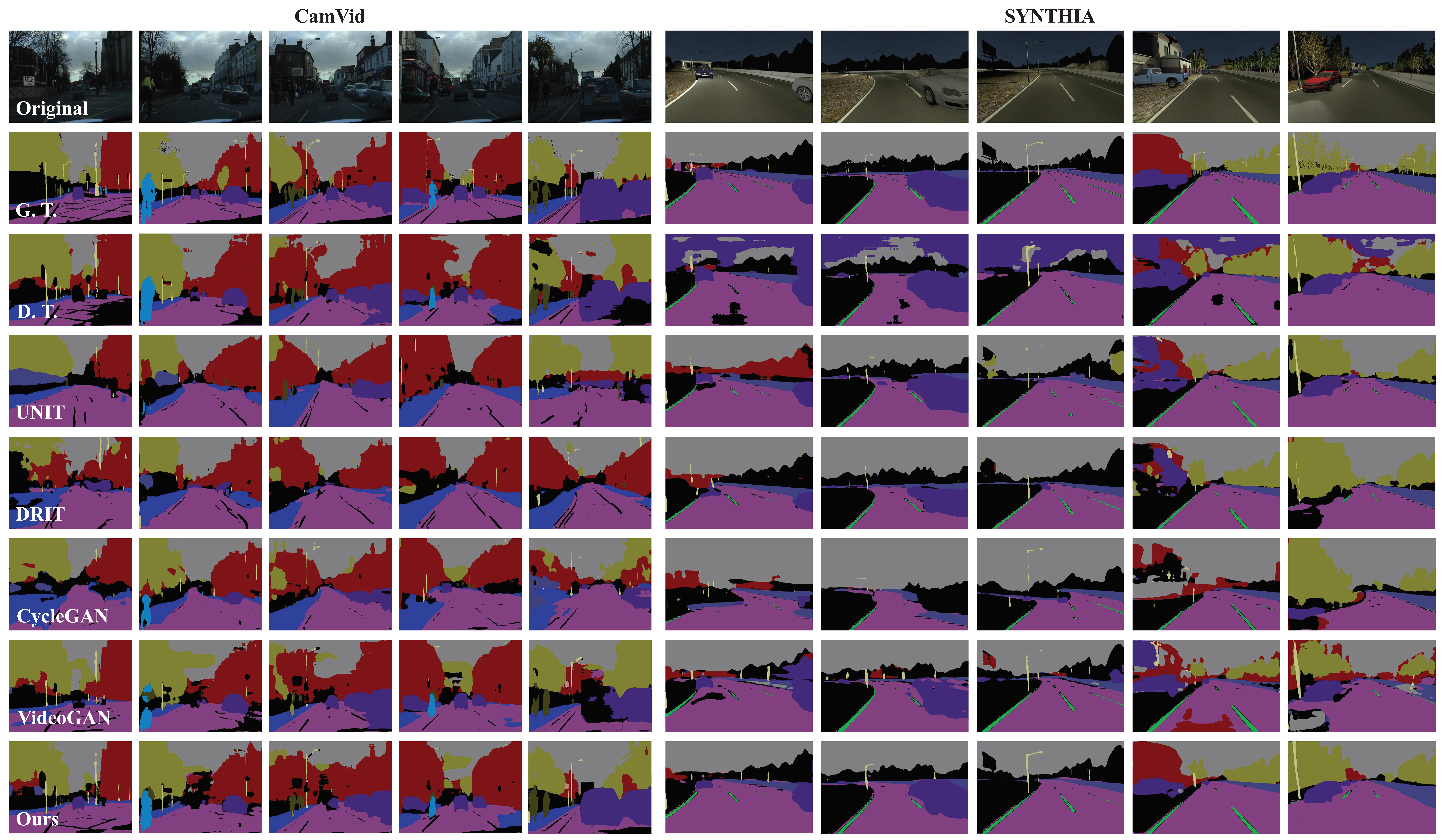}
      \caption{Comparison of segmentation results yielded by PSPNet using different I2I adaptation approaches. The original images, segmentation results using UNIT \cite{LiuMY01}, DRIT \cite{Lee01}, CycleGAN \cite{ZhuJ01}, VideoGAN \cite{Jiaweichen2020} and our Mutual-GAN are presented. Since we do not reproduce the OP-GAN \cite{Xie_2020_ECCV}, its segmentation results are excluded from the figure. (G. T.--Ground truth, D. T.--Direct transfer)}
      \label{fig:seg_comparison}
   \end{center}
   \vspace{-5mm}
\end{figure*}

\paragraph{\bf Visualization of segmentation results.}
The segmentation results yielded by PSPNet with different I2I adaptation approaches are presented in Fig.~\ref{fig:seg_comparison}. It can be observed that our Mutual-GAN can significantly improve the quality of semantic segmentation results of images under unfavorable weathers, which illustrates its potential for real applications.

\begin{figure}[!tb]
   \begin{center}
      \includegraphics[width=\linewidth]{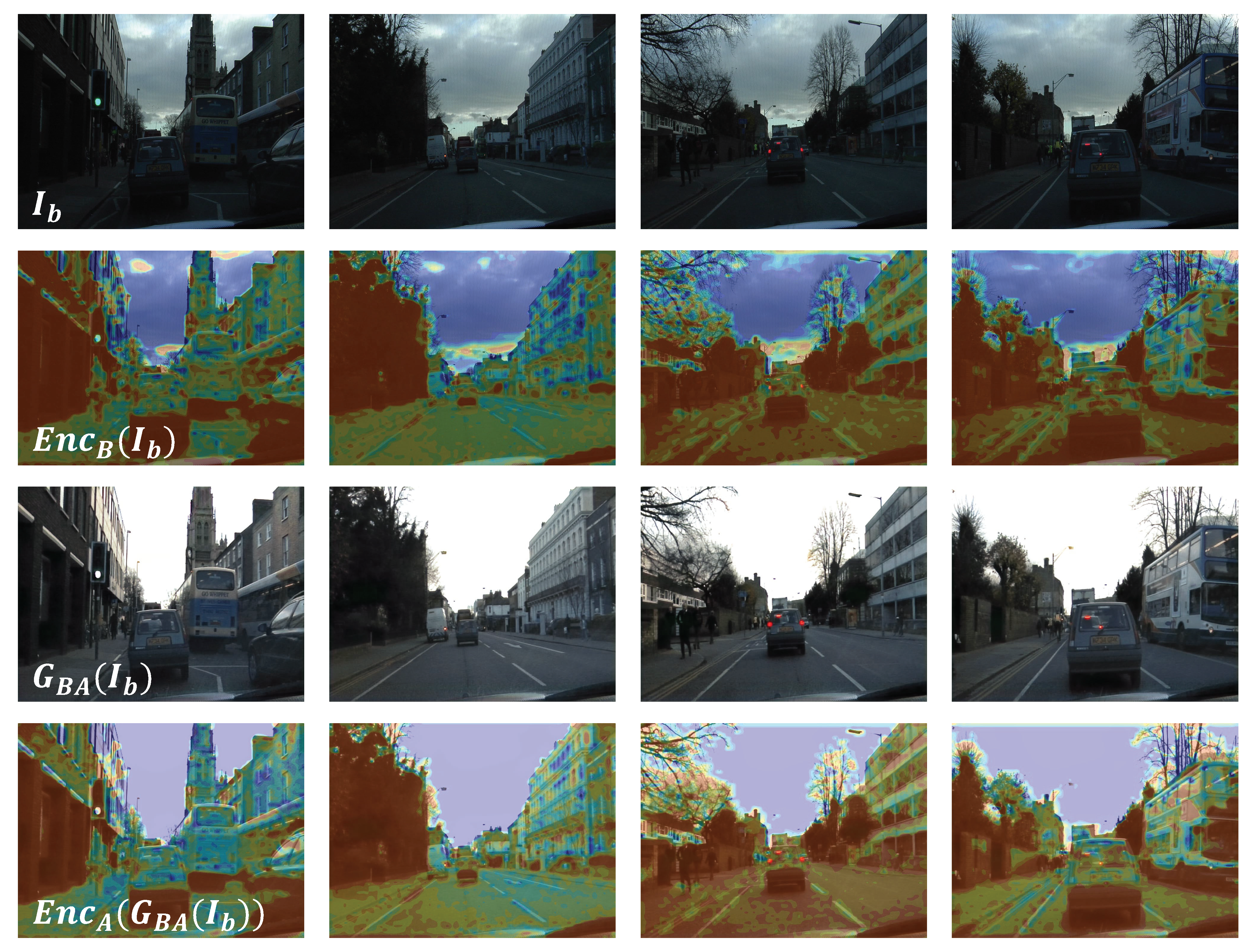}
      \caption{Attention maps generated by $Enc_B$ and $Enc_A$ for cloudy and translated images, respectively.}
      \label{fig:ablation}
   \end{center}
   \vspace{-5mm}
\end{figure}

\subsection{Ablation Study}
To assess the contribution made by each component of our Mutual-GAN (\emph{i.e.,} content feature encoder and mutual information constraint), we conduct an ablation study and present the evaluation results in this section.

\vspace{3mm}
\noindent{\bf Content feature encoder.} As illustrated in Fig.~\ref{fig:ablation}, we visualize the attention maps of cloudy and translated images produced by $Enc_B$ and $Enc_A$ to validate whether they successfully disentangle the content features from domain/weather information. The attention maps are overlapped to the cloudy and translated images for a better observation. It can be observed that content feature encoders ($Enc_B$ and $Enc_A$) activate the similar areas related to image-objects, such as buildings, trees and cars, and ignore those containing more domain/weather information, \emph{e.g.,} sky. Therefore, the preservation of image-objects can be achieved by narrowing down the distance between $Enc_B(I_b)$ ($z_b$) and $Enc_A(G_{BA}(I_b))$ ($z_{ba}$), \emph{i.e.,} $\mathcal{L}_{MI}(G_{BA}, Enc_A, Enc_B)$ defined in Eq.~\ref{eq:objective}.

\vspace{3mm}
\noindent{\bf Mutual information constraint.} To validate the effectiveness of our mutual information constraint, we conduct an ablation study on the two datasets. The evaluation results are presented in Table~\ref{table_camvid_quantity} and Table~\ref{table_synthia_quantity}. The mIoU of PSPNet remarkably degrades while testing on the images translated by Mutual-GAN without $\mathcal{L}_{MI}$---mIoU degrades to $41.82\%$ and $56.57\%$ on CamVid and SYNTHIA, respectively. The experimental results demonstrate that our mutual information constraint effectively enforces the Mutual-GAN to preserve the image-objects during cross-weather adaptation.

\section{Conclusion}
In this paper, we propose a novel generative adversarial network (namely Mutual-GAN) to alleviate the accuracy decline when daytime-trained neural network is applied to videos captured under adverse weather conditions. The proposed Mutual-GAN adopts mutual information constraint to preserve image-objects during cross-weather adaptation, which is an unsolved problem for most unsupervised image-to-image translation approaches (\emph{e.g.,} CycleGAN). The proposed Mutual-GAN is evaluated on two publicly available driving video datasets (\emph{i.e.,} CamVid and SYNTHIA). The experimental results demonstrate that our Mutual-GAN can yield visually plausible translated images and significantly improve the semantic segmentation accuracy of daytime-trained deep learning network while processing videos under challenging weathers.

\vspace{3mm}
\noindent{\bf Limitation \& Future work.} We notice the limitation of this study: the weathers are translated only under two scenarios, \emph{i.e.,} cloudy \emph{vs.} sunny and nighttime \emph{vs.} daytime, and plan to resolve it in our future work.

{\small
\bibliographystyle{ieee_fullname}
\bibliography{egbib_ssl}
}

\end{document}